\documentclass[11pt,a4paper]{article}
\usepackage{booktabs} 
\usepackage[hyperref]{acl2020}
\usepackage{times}
\usepackage{latexsym}
\usepackage{graphicx}
\usepackage{subcaption}

\usepackage{microtype}

\aclfinalcopy 

\title{ReINTEL Challenge 2020: A Multimodal Ensemble Model for Detecting Unreliable Information on Vietnamese SNS}

\author{Nguyen Manh Duc Tuan \\
  Toyo Unversity, Japan \\
  \texttt{ductuan024@gmail.com} \\\And
  Pham Quang Nhat Minh \\
  Aimesoft JSC, Vietnam \\
  \texttt{minhpham@aimesoft.com} \\}

\date{}

\begin{document}
\maketitle

\begin{abstract}
In this paper, we present our methods for unrealiable information identification task at VLSP 2020 ReINTEL Challenge. The task is to classify a piece of information into reliable or unreliable category. We propose a novel multimodal ensemble model which combines two multimodal models to solve the task. In each multimodal model, we combined feature representations acquired from three different data types: texts, images, and metadata. Multimodal features are derived from three neural networks  and fused for classification. Experimental results showed that our proposed multimodal ensemble model improved against single models in term of ROC AUC score. We obtained 0.9445 AUC score on the private test of the challenge.
\end{abstract}

\section{Introduction}
\label{sec:intro}

Recently, fake news detection have received much attention in both NLP and data mining research community. This year, for the first time, VLSP 2020 Evaluation Campaign organizers held ReINEL Challenge~\cite{le2020reintel} to encourage the development of algorithms and systems for detecting unreliable information on Vietnamese SNS. In ReINTEL Challenge 2020, we need to determine a piece of information containing texts, images, and metadata is reliable or unreliable. The task is formalized as a binary classification problem and training data with unreliable/reliable labels was provided by VLSP 2020 organizers.

In this paper, we present a novel multimodal ensemble model for identifying unreiable information on Vietnamese SNS. We use neural networks to obtain feature representations from different data types. Multimodal features are fused  and put into a sigmoid layer for classification. Specifically, we use BERT model to obtain feature representations from texts, a multi-layer perceptron to encode metadata and text-based features, and a fine-tuned VGG-19 network to obtain feature representations from images. We combined two single models in order to improve the accuracy of fake news detection. Our proposed model obtained 0.9445 ROC AUC score on the private test of the challenge.

\section{Related Work}
\label{sec:work}

Approaches to fake news detection can be roughly categorized into categorises: content-based methods, user-based methods and propagation-based methods.

In content-based methods, content-based features are extracted from textual aspects, such as from the contents of the posts or comments, and from visual aspects. Textual features can be automatically extracted by a deep neural network such as CNN ~\cite{kaliyar2020fndnet,tian2020stance}. We can manually design textual features from word clues, patterns, or other linguistic features of texts such as their writing styles~\cite{ghosh2018towards,Wang2018EANNEA,yang2018ticnn}. We can also analyze unreliable news based on the sentiment analysis ~\cite{8508256}. Furthermore, both textual and visual information can be used together to determine fake news by creating a multimodal model~\cite{zhou2020safe,Khattar2019,yang2018ticnn}.
 
We can detect fake news by analysing social network information including user-based features and network-based features. User-based features are extracted from user profiles~\cite{shu2019role,8424744,duan2020rmit}. For example, number of followers, number of friends, and registration ages are useful features to determine the credibility of a user post~\cite{Castillo2011}. Network-based features can be extracted from the propagation of posts or tweets on graphs~\cite{zhou2019networkbased,ma-etal-2018-rumor}.
 
\section{Methodology}
\label{sec:method}

In this section, we describe methods which we have tried to generate results on the test dataset of the challenge. We have tried three models in total and finally selected two best models for ensemble learning.

\subsection{Preprocessing}

In the pre-processing steps, we perform following steps before putting data into models.

\begin{itemize}
    \item We found that there are some emojis written in text format such as ``:)'', ``;)'', ``=]]'', ``:('', ``=['', etc. We converted those emojis into sentiment words ``happy'' and ``sad'' in Vietnamese respectively.
    
    \item We converted words and tokens that have been lengthened into short form. For example, ``Coooool'' into ``Cool'' or ``*****'' into ``**''. 
    \item Since many posts are related to COVID-19 information, we changed different terms about COVID-19 into one term, such as ``covid'', ``ncov'' and ``convid'' into ``covid'', for consistency.
    
    \item We used VnCoreNLP toolkit~\cite{vu-etal-2018-vncorenlp} for word segmentation.
\end{itemize}

Since meta-data of news contains a lot of missing values, we performed imputation on four original metadata features. We used the mean values to fill missing values for three features including the number of likes, the number of shares, and the number of comments. For the timestamp features, we applied the MICE imputation method~\cite{micearticle}.

We found that there are some words written in incorrect forms, such as 's.{\'a}th{\d{a}}i' instead of 's{\'a}t h{\d{a}}i'. One may try to convert those words into standard forms, but as we will discuss in Section~\ref{sec:analysis}, keeping the incorrect form words actually improved the accuracy of models.

We converted the timestamp feature into 5 new features: day, month, year, hour and weekday. In addition to metadata features provided in the data, we extracted some statistic information from texts: number of hashtags, number of urls, number of characters, number of words, number of question-marks and number of exclaim-marks. For each user, we counted the number of unreliable news and the number of reliable news that the user have made and the ratio between two numbers, to indicate the sharing behavior~\cite{shu2019role}. We also created a Boolean variable to indicate that a post contains images or not. In total, we got 17 features including metadata features. All the metadata-based features will be standardized by subtracting the mean and scaling to unit variance, except for the Boolean feature.

\begin{figure*}[!ht]
\begin{subfigure}{.5\textwidth}
  \centering
  \includegraphics[width=.8\linewidth]{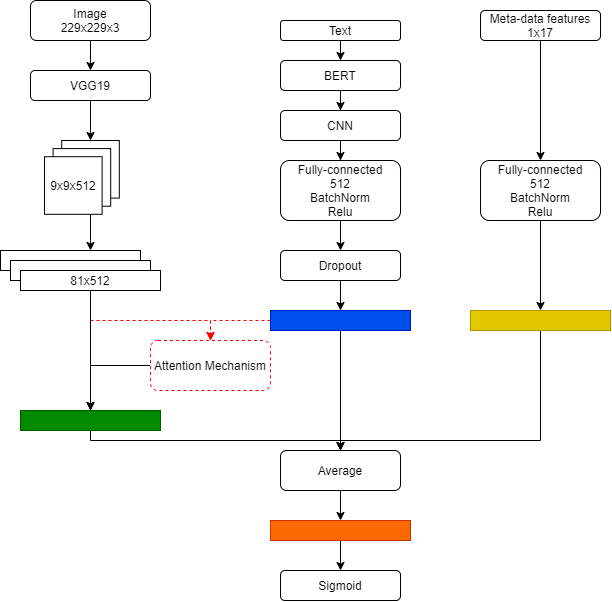}
  \caption{General structure for each model.}
  \label{fig:genral}
\end{subfigure}%
\begin{subfigure}{.5\textwidth}
  \centering
  \includegraphics[width=.8\linewidth]{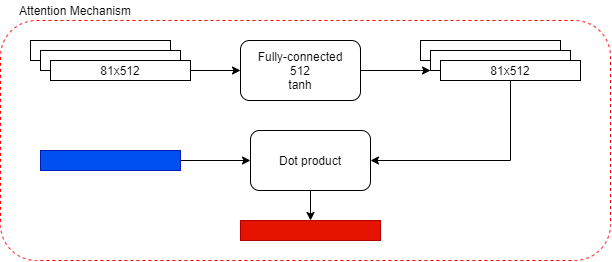}
  \caption{Attention Mechanism.}
  \label{fig:attention}
\end{subfigure}
\caption{General Model Architecture}
\label{fig:general_att}
\end{figure*}

\subsection{General Model Architecture}

Figure~\ref{fig:general_att} shows the general model architecture of three models we have tried. In all models, we applied the same strategy for image-based features and meta-data based features. For metadata-based features, we passed it into a fully-connected layer layer with batch normalization. We found that there are posts having one or more images and there are posts having have no image. For posts containing images, we randomly chose one image as the input. For other posts, we created a blank image as the input. We then fine-tuned VGG-19 model on the images of the training data. After that, we used the output prior the fully-connected layer as image-based features. Instead of taking averages of all vectors of pixels, we applied the attention mechanism as shown in Figure~\ref{fig:attention} to obtain the final representation of images.

\begin{figure*}[!ht]
\begin{subfigure}{.5\textwidth}
    \centering
    \includegraphics[width=0.6\linewidth]{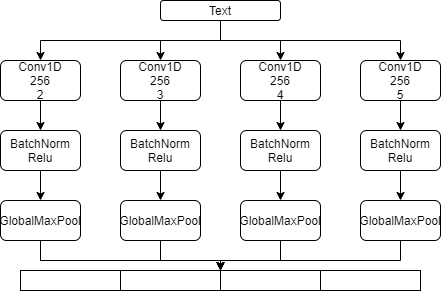}
    \caption{Model 1.}
    \label{fig:model1}
\end{subfigure}%
\begin{subfigure}{.5\textwidth}
    \centering
    \includegraphics[width=0.6\linewidth]{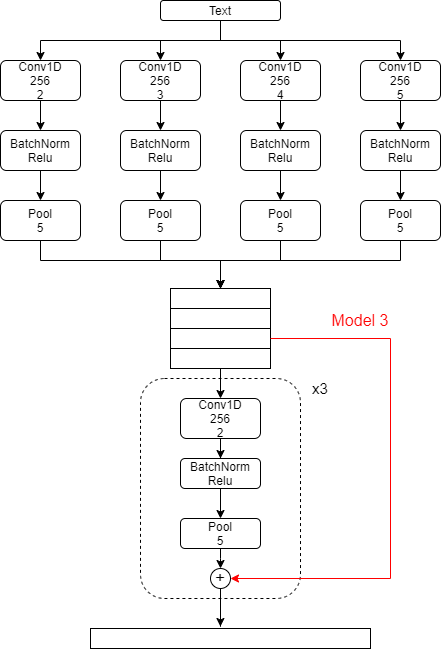}
    \caption{Model 2 and 3.}
    \label{fig:model2}
\end{subfigure}
\caption{Text-based features extractor for each model.}
\label{fig:fig}
\end{figure*}

In the following sections, we describe three variants that we made from the general architecture.

\subsection{Model 1}

In the first model (Figure \ref{fig:model1}), we obtained the embedding vector of a text using BERT model~\cite{devlin2019bert}. After that, we used 1D-CNN~\cite{kim2014convolutional} with filter sizes 2, 3, 4, and 5. By doing that, we can use more information from different sets of word vectors for prediction. We flattened and concatenated all the output from 1D-CNN and passed into a fully-connected layer with with a batch normalization layer. Finally, we took averages of features of texts, images and metadata and passed them into a sigmoid layer for classification.

\subsection{Model 2}

In the second model (Figure \ref{fig:model2}), there are some changes in comparison with the first model. After passing the embedding vectors through various layers of 1D-CNN, we stacked those outputs vertically and passed into three additional 1D-CNN layers. 

\subsection{Model 3}

In the third model (Figure \ref{fig:model2}), we just slightly changed the second model by adding a shortcut connections between input and the output of each 1D-CNN layer.

\subsection{Ensemble Model}

For the final model, we selected two best models among three above models and took averages of probabilities returned by the two models to obtain the final result.

\section{Experiments and Results}
\label{sec:analysis}

In experiments, we used the same parameters as showed in Table~\ref{table:1} for all proposed models. We reported ROC-AUC scores on the private test data.

In the first experiment, we compared two ways of preprocessing texts: 1) converting words in incorrect forms into corrected forms; and 2) keeping the incorrect forms of words. The text is put through PhoBERT~\cite{phobert} to get the embedded vectors. In this experiment, we did not apply the attention mechanism. Table~\ref{table:2} shows that keeping the original words obtained better ROC-AUC score.

Next, we compared the effects of two different pre-trained BERT models for Vietnamese: PhoBERT and Bert4news\footnote{Bert4News is available on: \url{https://huggingface.co/NlpHUST/bert4news}}. Table~\ref{table:3} shows that Bert4news model is significantly better than PhoBERT model. Furthermore, when we added the proposed attention mechanism to get feature representations for images, we obtained 0.940217 AUC score.

Table \ref{table:4} shows results for three models which we have described in section~\ref{sec:method}. We got 0.939215 with model 1, 0.919242 with model 2, and 0.940217 with model 3. The final model is derived from model 1 and model 3 by calculating the average of results returned by model 1 and model 3. We obtained 0.944949 of ROC-AUC using that simple ensemble model.

\begin{table}[!t]
  \centering
  \begin{tabular}{|p{3cm}|p{1cm}|}
  \hline
  \textbf{Hyper-parameter} & \textbf{Value}\\
  \hline
    FC layers & 512\\
  \hline
    Dropout & 0.2\\
  \hline
    Pooling size & 5\\
  \hline
    1D-Conv filters & 256\\
  \hline
    Learning parameter & 2e-5\\
  \hline
    Batch size & 16\\
  \hline
\end{tabular}
\caption{Parameters Setting}
\label{table:1}

\end{table}

\begin{table}
  \begin{tabular}{lc}
    \toprule
    \textbf{Run} & \textbf{ROC-AUC}\\
    \midrule
    Convert words to correct forms & 0.918298\\
    Keep words in incorrect forms & 0.920608\\
  \bottomrule
\end{tabular}
\caption{Two ways of preprocessing texts.}
\label{table:2}
\end{table}

\begin{table}[!t]
  \begin{tabular}{lcl}
    \toprule
    \textbf{Run} & \textbf{ROC-AUC}\\
    \midrule
    PhoBERT & 0.920608\\
    Bert4news & 0.927694\\
    Bert4news + attention & 0.940217\\
  \bottomrule
\end{tabular}
\caption{Comparison of different pre-trained models and using attention mechanism}
\label{table:3}
\end{table}

\begin{table}[!t]
  \centering
  \begin{tabular}{lcl}
    \toprule
    \textbf{Run} & \textbf{ROC-AUC}\\
    \midrule
    Model 1 & 0.939215\\
    Model 2 & 0.919242\\
    Model 3 & 0.940217\\
    Ensemble & 0.944949\\
  \bottomrule
\end{tabular}
\caption{Final results}
\label{table:4}
\end{table}

\section{Discussion}
\label{sec:lessons}

Since there may be more than one images in a post, we have tried to use one image as input or multiple images (4 images at most) as input. In preliminary experiments, we found that using only one image for each post obtained higher result in development set, so we decided to use one images in further experiments.

We have showed that keeping words in incorrect forms in the text better than fixing it to the correct forms. A possible explanation might be that those texts may contain violent contents or extreme words and users use that forms in order to bypass the social media sites' filtering function. Since those words can partly reflect the sentiment of the text, the classifier may gain benefit from it. The reason is that unreliable contents tend to use more subjective or extreme words to convey a particular perspective~\cite{8508256}.

We also showed that by using the proposed attention mechanism, the result improved significantly. This result indicates that images and texts are co-related. In our observation, images and texts of reliable news are often related while in many unreliable news, posters use images that do not relate to the content of the news for click-bait purpose.

We found that convolution layers are useful and textual features can be well extracted by CNN layers. \citealp{conneau2017deep} has showed that a deep stack of local operations can help the model to learn the high-level hierarchical representation of a sentence and increasing the depth leads to the improvement in performance. Also, deeper CNN with residual connections can help to avoid over-fitting and solves the vanishing gradient problem ~\cite{kaliyar2020fndnet}.

\section{Conclusion}
\label{sec:conclusion}
\subsection{Summary}

We have presented a multimodal ensemble model for unreliable information identification on Vietnamese SNS. We combined two neural network models which fuse multimodal features from three data types including texts, images, and metadata. Experimental results confirmed the effectiveness of our methods in the task.

\subsection{Future work}

As future work, we plan to use auxiliary data to verify if a piece of information is unreliable or not. We believe that the natural way to make a judgement in fake news detection task is to compare a piece of information with different information sources to find out relevant evidences of fake news.

\bibliography{vlsp2020}
\bibliographystyle{acl_natbib}

\end{document}